# CausalLP: Learning causal relations with weighted knowledge graph link prediction

Utkarshani Jaimini
Artificial Intelligence Institute,
University of South Carolina
Columbia, USA
ujaimini@email.sc.edu

Cory Henson
Bosch Center for Artificial
Intelligence
Pittsburgh, USA
cory.henson@us.bosch.com

Amit Sheth
Artificial Intelligence Institute,
University of South Carolina
Columbia, USA
amit@sc.edu

## ABSTRACT
Causal networks are useful in a wide variety of applications, from medical diagnosis to root-cause analysis in manufacturing. In practice, however, causal networks are often incomplete with missing causal relations. This paper presents a novel approach, called CausalLP, that formulates the issue of incomplete causal networks as a knowledge graph completion problem. More specifically, the task of finding new causal relations in an incomplete causal network is mapped to the task of knowledge graph link prediction. The use of knowledge graphs to represent causal relations enables the integration of external domain knowledge; and as an added complexity, the causal relations have weights representing the strength of the causal association between entities in the knowledge graph. Two primary tasks are supported by CausalLP: causal explanation and causal prediction. An evaluation of this approach uses a benchmark dataset of simulated videos for causal reasoning, CLEVRER-Humans, and compares the performance of multiple knowledge graph embedding algorithms. Two distinct dataset splitting approaches are used for evaluation: (1) random-based split, which is the method typically employed to evaluate link prediction algorithms, and (2) Markov-based split, a novel data split technique that utilizes the Markovian property of causal relations. Results show that using weighted causal relations improves causal link prediction over the baseline without weighted relations.

## 1 INTRODUCTION
Causal networks are structured as a directed, acyclic graph with edges representing the causal links between entities. Each causal link may be annotated with weights representing the strength of the causal association. Traditional techniques for generating causal networks rely solely on the use of observation data with datasets that are often incomplete and lack important information about the underlying causal structures, leading to an incomplete causal network. If the incomplete causal network is encoded as a knowledge graph (KG), then the task of finding missing causal relations can be formulated as a knowledge graph completion problem, i.e. finding missing links in the knowledge graph.

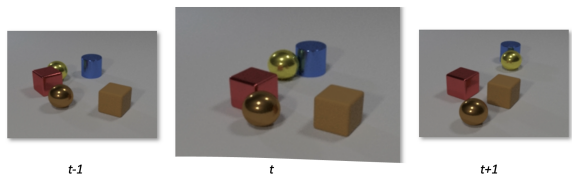

Figure 1: A snapshot of collision events in a video at time *t-1*, *t*, and *t+1*. There are four consecutive collision events that occur: 1) the red cube enters from the left, 2) the red cube collides with the yellow ball, 3) the yellow ball hits the blue cylinder, and 4) the blue cylinder moves.

In this paper, we present an approach to finding new causal relations using KG link prediction methods. This approach is composed of four primary phases: (1) encoding known causal relations into a causal network, (2) translating the causal network into a knowledge graph, (3) learning knowledge graph embedding for the causal relations, and (4) predicting new causal links in the knowledge graph.

In a knowledge graph, causal links may be encoded as follows: *<cause-entity, causes, effect-entity, w>* with the **causes** relation linking the cause and effect entities. Each causal link is associated with a *causal weight, w*, that represents the causal influence of the *cause-entity* on the *effect-entity*. This causal influence is measured by performing an intervention on the *cause-entity* and observing its outcome on the *effect-entity* [12]. In the next phase, a KG embedding (KGE) model is learned. Any KGE algorithm may be used for this task. However, current algorithms do not incorporate relation weights into the learned embedding model. To overcome this issue, FocusE is used to assimilate the *causal weights* into the KGE model [10]. In the final phase, KG link prediction is used to find new causal links. More specifically, two tasks are performed: causal explanation and causal prediction. When implemented with link prediction, causal explanation is mapped to the task of finding the type of the head (i.e. *cause-entity*) of a causal link, and causal prediction is mapped to the task of finding the type of the tail (i.e. *effect-entity*) of a causal link.

Consider an example shown in Figure 1. The events occurring in the video frames can be encoded in a causal KG. At *t-1* *<the red cube enters from the left, **causes**, the red cube collides with the yellow ball>*. This subsequently leads to *t* where *<the red cube collides with the yellow ball, **causes**, the yellow ball hit with the blue cylinder>*. Eventually this leads to *t+1* where *<the yellow ball hits the blue cylinder, **causes**, the blue cylinder to move>*. Now consider the following causal explanation query: Explain the cause of the event *the red cube colliding with the yellow ball* which occurs at *t*. The answer would be the prior event *the red cube enters from the left* which occurs at *t-1*. Similarly, consider the causal prediction query: Predict the effect of the event *the red cube colliding with the yellow ball* which occurs at *t*. The answer would be the subsequent event *the blue cylinder moves* which occurs at *t+1*. From these examples, we can see that the answer to causal explanation queries requires predicting a causal link to prior events, and the answer to a causal prediction queries requires predicting a causal link to subsequent events.

With the traditional approach to evaluating KG embedding algorithms, links are randomly split into a train and test set. In the case of a causal KG, such an approach could lead to model bias. This is due to the fact that there may be multiple causal relations connecting a cause and effect entity in the KG. To resolve this issue we propose a novel approach to splitting the data, called Markov-based split, that is based on the local Markov property of the causal links.

Due to this split, it may be the case for causal explanation, at *t*, that some of the causal links from prior frames (i.e. *t-1*) could be in train set. Similarly, for causal prediction at *t*, some of the latter links (i.e. *t+1*) could be in train set. During the evaluation this could lead to model bias. To resolve this issue we propose to split the data based on the local Markov property of the causal links (i.e. Markov-based splitting). The causal links in *t* are independent of causal links in *t+1* given the causal links in *t-1*. For causal explanation in frame *t*, the causal links with events spanning from *t-1* to *t* are in the test set. Similarly for causal prediction in frame *t*, the causal links with events spanning from *t* following *t+1* are in the test set.

For the causal prediction task, given a *cause-entity* we want to predict the type of the *effect-entity*. There are several causal links that may provide relevant information: causes, causedBy, and causesType. To make a causal prediction in this case, the causesType link which links the *cause-entity* to the type of *effect-entity* would need to be assigned to the test set (see Figure 3). But with a random split, the other causal relations (causes and causedBy) connecting these entities may be assigned to the training set. In a real-world scenario, we would not know any of this additional causal information when inferring the effect of an event. Thus, by providing this information during training would lead to bias in the prediction model and artificially inflate the prediction performance results.

The proposed approach is evaluated using a benchmark dataset for causal reasoning, CLEVRER-Humans [9]. The dataset contains a collection of videos showing objects colliding in a simulated environment. Each video in the dataset is annotated with causal relations between events, along with the associated causal weights. Both causal link prediction tasks are evaluated with multiple common KGE algorithms, including DistMult, HolE, TransE, and ComplEx. Results of the evaluation show that the addition of *causal weights* leads to improved performance for all algorithms, and also when using either the random-split and Markov-based split.

The main contributions of this paper include:

(1) A novel formulation of the task of finding missing causal relations in an incomplete causal network as a KG completion problem.

(2) Demonstration of the approach for causal link prediction using a benchmark dataset of video simulation data.
(3) Evaluation of the approach using multiple KG embedding algorithms, which shows that incorporating causal weights leads to improved performance.
(4) Definition and use of a novel method, Markov-based data split, for evaluating the causal link prediction tasks.

The rest of the paper proceeds as follows: Section 2 describes the related work. The problem formulation is defined in Section 3 followed by the methodology in Section 4. Section 5 details the evaluation with the results and discussion outlined in Section 6. Section 7 provides a conclusion with future direction.

## 2 RELATED WORK

**Causal discovery:** Causal discovery algorithms fall under two major categories: constraint based and score based [2]. The constraint based methods use conditional independence relations in the observational data to find Markov equivalence classes of directed causal structures. The score based methods use structural equation models to find unique causal structures under certain assumptions. Another approach to causal discovery, knowledge guided greedy score based approach, uses prior knowledge about the causal structure (knowledge about the edges, i.e. presence or absence of a directed or an un-directed edge) between entities and observational data to learn causal graphs [3]. The research demonstrates that the prior structural knowledge improves causal graph discovery. The focus of the above research, however, is on causal graphs and not knowledge graphs.

**Causal knowledge graphs:** CauseNet is a knowledge graph of causal concepts and relations extracted from semi and unstructured web sources [5]. The CauseNet links are of the form <subject may-Cause object>. Linguistic patterns are used to extract the causal statements. Statements with words such as "cause(d)", "result(ed)", "lead(s)/led", and "associated with" are termed as causal statements. The causal concepts, i.e. cause and effect, in the statements are presumed to be nouns similar to ConceptNet and WordNet. The linguistic patterns extract explicit causal relations from statements, but does not capture implicit mentions of causal relation and causal concepts spanning across sentences. Since the causal relations are extracted from web, they suffer from societal bias in the data. Some of the extracted biased causal relations include "Autism is caused by vaccination" and "HIV is caused by homosexuality" [5].

**Causal link prediction:** The existing techniques for KG link prediction are used to predict general links and are not tailored for predicting causal links specifically. However, [4] generates an event-related causal knowledge graph from Wikipedia articles and Wikidata with causal predicates, hasCause and hasEffect, associated with a given event. The nodes in the graph are events and the edges represent cause-effect relations. It aims to predict future events by analyzing the underlying causes and effects of similar past events. Existing KG link prediction techniques are used to evaluate the causal relation prediction task.

This paper focuses on finding missing causal relations using knowledge graph completion. The CausalLP approach proposed in this paper predicts new causal links in a knowledge graph utilizing four causal relations i.e. causes, causedBy, causesType, and causedByType. The approach uses the prior causal structure knowledge encoded in a causal network which in turn is represented in the causal knowledge graph[6].

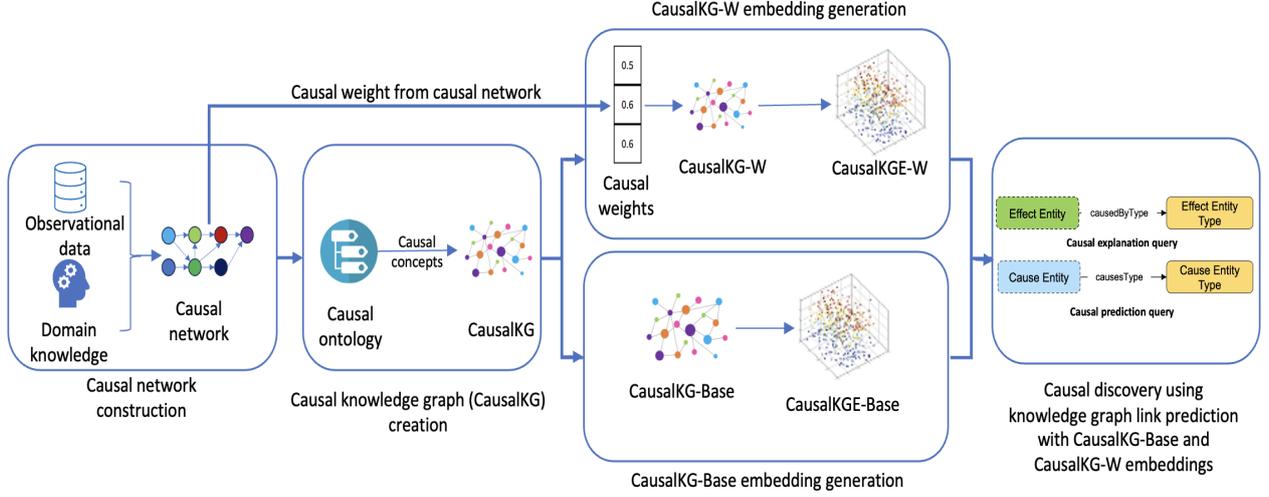

Figure 2: CausalLP has four primary phases: 1) encoding the causal associations in data as a causal network, 2) translating the causal network into a causal knowledge graph, 3) learning knowledge graph embeddings from the causal knowledge graph, and 4) using the knowledge graph embeddings for causal link prediction tasks.

## 3 PROBLEM FORMULATION

This section formulates the problem of finding missing causal relations as a KG link prediction task and defines the primary concepts, including causal relation, causal link, causal entity, causal weight, and causal knowledge graph.

**Causal knowledge graph:** A causal knowledge graph $CausalKG$ is a KG that includes causal knowledge in the form of causal entities, causal relations, and causal weights. $CausalKG = (N, R, E, E_c, W_c)$:

- $N$: a set of nodes representing entities
- $R$: a set of labels representing relations
- $E \subseteq N \times R \times N$: a set of edges representing links between pairs of entities. Each link has the form $<h, r, t>$, where $h$ is the head entity, $r$ is the relation, $t$ is the tail entity.
- $N_c \subseteq N$: a set of nodes representing causal entities
- $R_c \subseteq R$: a set of labels representing causal relations
- $W_c \subseteq \mathbb{R}$: a set of real numbers representing causal weights
- $E_c \subseteq N_c \times R_c \times N_c \times W_c$: a set of edges representing causal links connecting pairs of causal entities. Each causal link is a quad $<h_c, r_c, t_c, w_c>$, where $h_c$ is the head causal entity, $r_c$ is the causal relation, $t_c$ is the tail causal entity, and $w_c$ is the causal weight.

**Causal entity:** A causal entity $n_c \in N_c$ is an entity that is the head or tail of a causal link. There are two types of causal entities: cause-entity ($n_{cause}$) and effect-entity ($n_{effect}$) such that the cause-entity causes the effect-entity.

**Causal relation:** A causal relation $r_c \in R_c$ is a relation representing a causal association between entities. There are four types of causal relations:

- causes ($r_{causes} \in R_c$) is a causal relation from the cause-entity to the effect-entity.
- causedBy ($r_{causedBy} \in R_c$) is a causal relation from the effect-entity to the cause-entity; i.e. the inverse of causes.
- causesType ($r_{causesType} \in R_c$) is a causal relation from the cause-entity to the type of the effect-entity.
- causedByType ($r_{causedByType} \in R_c$) is a causal relation from the effect-entity to the type of the cause-entity.

**Causal weight:** A causal weight $w \in W_c \subseteq \mathbb{R}$ is a real number associated with a causal link. It quantifies the responsibility or contribution of the cause-entity in causing the effect-entity.

**Causal link:** A causal link $e_c \in E_c$ is an edge in the causal KG connecting a pair of causal entities with a causal relation and an associated causal weight. Causal link is a quad $<h_c, r_c, t_c, w_c>$, where $h_c$ is the head causal entity, $r_c$ is the causal relation, $t_c$ is the tail causal entity, and $w_c$ is the causal weight.

**Causal link prediction:** Causal link prediction is the task of finding new causal links in a CausalKG. Given a CausalKG G, this task can be implemented using knowledge graph link prediction. There are two distinct methods for finding causal links: causal prediction and causal explanation.

(1) Causal prediction: given a cause-entity ($n_{cause} \in N_c$) and the $causesType$ relation ($r_{causesType} \in R_c$), find the type ($t$) of the associated effect-entity such that $<n_{cause}, r_{causesType}, t, w_c> \in G$ holds.
(2) Causal explanation: given an effect-entity ($n_{effect} \in N_c$) and the $causedByType$ relation ($r_{causedByType} \in R_c$), find the type ($t$) of the associated cause-entity such that $<n_{effect}, r_{causedByType}, t, w_c> \in G$ holds.

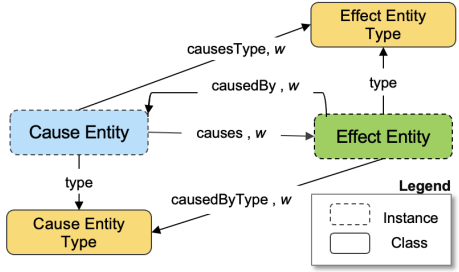

**Figure 3: Reified causal relation, *causesType* and *causedBy-Type*. The *causesType* is a reified relation from a cause-entity instance to the type of an effect-entity. The *causedByType* is a reified relation from an effect-entity instance to the type of a cause-entity.**

## 4 METHODOLOGY

The proposed approach is structured into four primary phases (see figure 2): (1) finding and encoding the known causal relations into a causal network, (2) translating the causal network into a CausalKG, conformant to the causal ontology [6], (3) learning KG embeddings for the CausalKG, and (4) predicting new causal links in the KG.

### 4.1 Causal Network

A causal network is a graphical model that describes the cause-and-effect relationships between the nodes and is represented as a causal Bayesian network. It is a directed acyclic graph where the nodes of the network denote events and the edges represent the causal association between them. $CN = (N^{cn}, E^{cn}, W^{cn})$, such that $N^{cn}$ is the set of nodes in the causal network, $E^{cn}$ is the set of edges between nodes, and $W^{cn}$ is the set of causal weights associated with the edges. The direction of the edge denotes the direction of the causal association. Each edge has a causal weight, $w \in W$, which measure the strength of the edge between the nodes. The causal weight represents the total causal effect estimated using do-calculus. The total causal effect is the measure of the strength of the change of a given node on its direct linked node [11, 12]. Given an edge, $e \in E^{cn}$, between two nodes ($n_1 \in N^{cn}, n_2 \in N^{cn}$), the total causal effect can be estimated as an expected value ($EV$) of intervention on $n_1$ using do-calculus, $EV[n_2|do(n_1)]$. The causal network satisfies the local Markov property where given the direct causes of a node, it is independent of its non-effects [8, 12].

### 4.2 Causal Knowledge Graph

The task of translating information from a causal network into a causal KG is fairly straightforward:

- $N^{cn} \rightarrow N_c$: nodes in the causal network become causal entities in the CausalKG
- $W^{cn} \rightarrow W_c$: weights in the causal network become causal weights in the CausalKG
- $E^{cn} \rightarrow E_c$: edges in the causal network become causal links in the CausalKG, of the form $<n_{cause}, r_{causes}, n_{effect}, w_c>$

Additional causal links are added to the KG as appropriate, including those utilizing the other causal relations: *causedBy*, *causesType*, and *causedByType*. The resulting CausalKG contains all the information from the causal network and is conformant to the causal ontology [6, 7].

The causal ontology used for this task defines the structure and semantics of causal relations using the concepts grounded in causal AI, i.e causal Bayesian network and do-calculus [6]. More specifically, the ontology defines the primary concepts used to structure a CausalKG, including causal entities, causal relations, and causal weights.

The CausalKG is used for finding missing causal links using KG link prediction. For the task of causal explanation, the goal is to predict the type of a cause-entity that's linked to an effect-entity and not to predict the specific cause-entity instance. However, the effect-entity does not link directly with the cause-entity type. Rather it is connected through a two-hop path: $<n_{effect}, r_{causedBy}, n_{cause}>, <n_{cause}, rdf:type, type>$. KG link prediction models can only make predictions about directly linked entities. To overcome this issue, CausalKG uses a reified relation *causedByType* ($r_{causedByType} \in R_c$) to add links connecting an effect-entity with the type of a cause-entity (see Figure 3). The same issue arises for the task of causal prediction where the goal is to predict the type of an effect-entity that's linked to a cause-entity. To overcome this issue, CausalKG uses a reified relation *causesType* ($r_{causesType} \in R_c$) to add links connecting a cause-entity with the type of an effect-entity.

The CausalKG can also integrate additional domain knowledge associated with the causal entities which are not explicitly mentioned in the causal network.

### 4.3 CausalKG Embedding and Link Prediction

CausalKG can be converted into a low-dimensional continuous latent vector space representation, called KG embeddings (KGE). The KG embeddings can then be used for downstream tasks such as link prediction, triple classification, entity classification, relation extraction, etc. [14]. CausalLP uses KG embedding algorithms to generate embeddings that will be used for finding causal links. The proposed approach learns two types of KGEs for a CausalKG: 1) CausalKGE-Base, embeddings without causal weights, and 2) CausalKGE-W, embeddings with causal weights.

The CausalKG-W uses causal weight to generate weighted KGEs [10]. The CausalKG embeddings for both CausalKGE-Base and CausalKGE-W are generated using KG embedding algorithms available in the Ampligraph library[1]. The CausalKGE-Base embedding is trained using the causal links, but ignoring the causal weights associated with each link. The CausalKGE-W embedding, on the other hand, is trained using the causal links with the causal weights. The links with high causal weight will have a high probability of being true. A link with low causal weight signifies an unlikely link, and links with causal weight zero are considered as negative samples. During the training, causal weights are used to update the output of the scoring layer of a KGE algorithm before feeding the scores to the loss layer [10]. The scores from the scoring layer are modulated based on the causal weight values associated with the links to obtain "weighted" scores. The CausalKGE-Base and CausalKGE-W embeddings are evaluated on the task of causal link prediction using KG link prediction techniques.

---

[1] https://docs.ampligraph.org/en/latest/

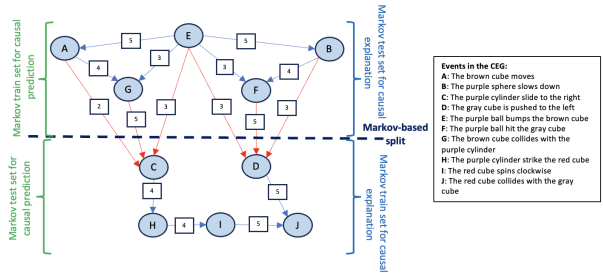

Figure 4: Causal event graph. The nodes in the graph represent events in a video and the edges represent causal relations. The edge label is a human annotated score symbolizing the strength of the causal relation. For example, the edge from event E to A represents the fact that event E causes event A. The edge label of 5 represents that event E is *extremely responsible* for causing event A. The horizontal dotted line exemplifies the Markov-based split.

The proposed approach, CausalLP, formalizes the problem of finding missing causal links as a KG link prediction task. The trained CausalKG embedding models, i.e. CausalKGE-Base and CausalKGE-W, are used to predict missing causal links between causal entities in the KG. More specifically, CausalLP is used for the task of causal explanation and causal prediction. Causal explanation aims to predict the cause of an effect and causal prediction aims to predict the effect of a cause. For a given causal link, causal explanation predicts links of form $<n_{effect}, r_{causedByType}, ?, w>$, and causal prediction predicts links of form $<n_{cause}, r_{causesType}, ?, w>$.

Given any dataset, with causal relations, causal entities, and weights associated with the causal links between the entities, CausalDisco can be used to generate a CausalKG and KG embeddings. The generated KG embeddings can then be used for causal relation discovery in the form of causal explanation and causal prediction. In the next section, we demonstrate and evaluate our approach using CLEVRER-Humans, a causal reasoning benchmark dataset [9].

## 5 EVALUATION

The proposed approach is evaluated using KG link prediction task for 1) causal explanation, given an effect-entity predict the type of the cause-entity of the causal link of form $<n_{effect}, r_{causedByType}, ?, w>$ and 2) causal prediction, given a cause-entity predict the type of effect-entity of the causal triple of form $<n_{cause}, r_{causesType}, ?, w>$ (see Figure 3). The above experiment is demonstrated using a benchmark dataset for causal reasoning, CLEVRER-Humans [9]. This section describes the CLEVRER-Humans dataset, data pre-processing steps, creation of a CausalKG from the dataset, experimental set up, evaluation metrics, and description of the evaluation for different CausalKG variations.

### 5.1 Data

CLEVRER-Humans is a causal reasoning benchmark dataset with human annotated causal judgement regarding physical events occurring in videos [9]. The dataset is based on the videos from CLEVRER, a simulated dataset of collision events for video representation and reasoning [16]. The videos consist of moving objects that are distinct in their shape (sphere, cube, and cylinder), color (blue, red, yellow, green, purple, gray, cyan and brown) and material (metal and rubber). Each object can participate in 27 distinct events such as enter, exit, collide, move, hit, bump, roll etc. Figure 1 shows an example snapshot of events occurring in a CLEVRER video. CLEVRER-Humans encodes the causal information from these events in the form of Causal Event Graph (CEG), where the nodes of the graph are descriptions of events in the videos and the directed edges between the nodes represent the causal links (Figure 4). The edges of the CEGs are scored by human annotators to determine the the strength of causal links between the nodes. The edges are scored between 1-5, such that 1 = *not responsible at all*, 2 = *a bit responsible*, 3 = *moderately responsible*, 4 = *quite responsible*, and 5 = *extremely responsible*.

### 5.2 Pre-processing the Data

The first step towards generation of a CLEVRER-Humans CausalKG involves pre-processing the CEGs.

*5.2.1 Structure of CEG.* The CEGs are considered as a proxy for a causal network. The pre-processing of the CEGs are done to ensure that they are consistent with the definition of causal network. The edges in the causal network represents causal links between the nodes. As the first step, we remove the edges with score 1, as it is determined that there is no causal responsibility between the two given nodes. Next, since a causal network is a directed acyclic graph, the edges responsible for cycles in the CEGs are removed. Finally, we remove the CEGs which 1) do not have any remaining causal links between the nodes, or 2) have depth < 2 from the root node to the leaf node, in order to satisfy the requirement for our Markov-based split. After pre-processing we are left with 764 CEGs.

*5.2.2 Event extraction.* The CLEVRER-Humans dataset contains 27 distinct events such as collide, enter, exit, halt, go, etc. We subdivided the events into two categories: binary and singular event. A binary event involves two participating objects, including events such as collide, bump, hit, bounce, sideswipe, etc. A singular event involves only a single participating object, including events such as enter, exit, stop, etc. Information about the event type and participating objects are extracted from the node descriptions in the CEG. This is accomplished by parsing the CEG JSON files provided by the dataset. We use the Berkeley neural semantic parser[2] and NLTK[3] stem lemmatizer to capture the root form of the event label, such as collide, hit, push, etc. instead of collided, hits, pushed, etc. The nodes that include a composition of multiple events, such as *The red ball collides with the blue sphere and hits the yellow cylinder* which consists of two events (i.e. collide and hit), are removed from the CEG. We are only interested in nodes that describe a single event.

*5.2.3 Object and object property extraction.* Along with the event type, we also extract the participating objects and their characteristics, such as color, shape, and material. There are some object characteristics that are mislabelled in the dataset, such as labeling an object color as gold rather than yellow. These mislabelling issues are identified and the terms are normalized.

---
[2] https://pypi.org/project/benepar/
[3] https://www.nltk.org/index.html

| Triple associated with Node A | Triple associated with Node G | Triple associated with Node C |
|---|---|---|
| < A type Move ><br>< A hasParticipant Brown cube > | < G type Collide ><br>< G hasParticipant Brown cube ><br>< G hasParticipant Purple cylinder > | < C type Slide ><br>< C hasParticipant Purple cylinder > |
| < A causedBy E ><br>< A causes C ><br>< A causes G > | < G causedBy E ><br>< G causedBy A ><br>< G causes C > | < C causedBy A ><br>< C causedBy E ><br>< C causedBy G ><br>< C causes H > |
| < A causesType Slide ><br>< A causesType Collide ><br>< A causedByType Bump > | < G causesType Slide ><br>< G causedByType Move ><br>< G causedByType Bump > | < C causesType Strike ><br>< C causedByType Move ><br>< C causedByType Collide ><br>< C causedByType Bump > |

**Figure 5: A snapshot of links in CLEVRER-Humans CausalKG derived from nodes A, G, and C in the CEG illustrated in Figure 4.**

### 5.3 CausalKG from CLEVRER-Humans

A CausalKG is generated from CLEVRER-Humans by encoding the causal information within the CEGs in RDF[4] format, conformant with the causal ontology. In addition to causal relations, the KG contains information about events (such as hit, collide, push, etc.) along with the participating objects and their characteristics. The CEGs are graphical representations of events in the videos. We use three ontologies to represent information from the CEGs: causal ontology, scene ontology (prefix *so:*), and semantic sensor network ontology (prefix *ssn:*). The causal ontology is used to represent the events (as causal entities), causal relations, and their associated causal weights (i.e. edge score). The scene ontology and sensor ontology are used to represent the additional information about the video, including scenes, objects, and object characteristics [13, 15]. Each video is represented as a scene (*so:Scene*) using concepts from the scene ontology. This includes representing and linking the events included in the scene (with the *so:includes* relation), the participating objects (with the *so:hasParticipant* relation), and the object characteristics (with the *ssn:hasProperty* relation) [15]. In total, the CausalKG from CLEVRER-Humans contains >48K links, 5664 entities, 31 entity types, and 10 relations.

### 5.4 Splitting the Data

We introduce a novel dataset splitting approach, Markov-based split, grounded in the local Markov property of causal relations. For the evaluation, we utilise two different techniques for splitting the data into train and test sets: 1) random data split and 2) Markov-based data split. In the random data split, the links in the CausalKG are randomly split into train, test, and validation sets following the 80:10:10 split ratio. Depending on which causal links are selected for training and testing, this approach could lead to model bias. For example, given the CEG shown in Figure 4, and considering the causal prediction query for event G (i.e. predict the type of effect-entity), there are several causal links that may provide relevant causal information (see Figure 5). Obviously the link <$G, causesType, Slide$> should not be included in training since this link to be predicted. Other relevant causal links may also lead to model bias, such as <$G, causes, C$> and <$C, causedBy, G$>, and should not be in the training set. In general, any causal link in the CausalKG that spans across the Markov-based split line should not be used for training in order to minimize model bias which leads to inflated model performance [1].

To mitigate the above issue, we introduce an additional pre-processing step using the Markov-based data split before generating a CausalKG. With the Markov-based data split, the initial train and test sets contains 80% and 20% of the total CEGs, respectively. From the 764 CEGs, 612 are in train and 152 are in test set. Next, the CEGs in the test set are further split at depth 1 from the root node, as illustrated by the horizontal dotted line in figure 4. The split is based on the local Markov property of a causal network; i.e. for a given direct cause of a node, it is independent of its non-effects. The nodes and edges on either side of the horizontal dotted line are denoted as markov-train and markov-test sets, depending on the task, i.e. either causal explanation or causal prediction. For the task of causal prediction, the nodes and relations in the upper half of the Markov-based split are included in the Markov-train set and the lower half in the Markov-test set, and vice-versa for causal explanation. Furthermore, causal links spanning across the horizontal dotted line (represented as red-colored edges in figure 4) are masked and moved to the Markov-test set. The CEGs in the train set, along with the nodes and relations in Markov-train set, are used to generate the CausalKG for CLEVRER-Humans which is then subsequently used for training the KG embeddings. The nodes and relations in the Markov-test are used to generate test links for evaluating the KG embeddings. The respective data splits are fed to the KGE algorithms to generate both CausalKGE-Base and CausalKGE-W embeddings which will be used to for the link prediction task for causal explanation and prediction.

### 5.5 Diversifying the Available Knowledge

The CausalKGE-Base and CasualKGE-W embeddings are generated and evaluated on different CLEVRER-Humans CausalKG subgraph structures for the causal explanation and causal prediction tasks, as shown in Figure 6. Various graph structures are used in order to evaluate how CausalLP performs when different types of information are available in the CausalKG. Specifically, there are three distinct sub-graph structures defined with an increasing level of expressivity. The first graph structure **C**, shown in Figure 6(a), contains only links with causal relations. The second graph structure **CT**, shown in Figure 6(b), contains links with causal relations and causal entity types (i.e. rdf:type). The third graph structure **CTP**, shown in Figure 6(c), contains links with causal relations, causal entity types, and objects related to causal entities (i.e. hasParticipant). We optimized the hyper-parameters for each of the above graph structures for both causal explanation and prediction. The CausalKGE models are trained on their respective optimized hyper-parameters. The optimized hyper-parameter can be found here [5]. The trained CausalKGEs are then used for causal link prediction tasks using well established KG link prediction methods.

### 5.6 Evaluation Metrics

CausalLP is evaluated by following the KG link prediction experiment design. Given the set of causal links $E_c$ in CausalKG, a set of

---

[4] https://www.w3.org/RDF/

[5] https://github.com/CausalKG/CausalLP/tree/main

CausalLP: Learning causal relations with weighted knowledge graph link prediction corrupted links $\mathcal{T}'$ are generated by replacing the head $h_c$ or tail $t_c$ of a set of causal links, $<h_c, r_c, t_c, w_c>$, with another causal entity in the KG. Replacing the head with $h'_c \neq h_c$ results in $<h'_c, r_c, t_c, w_c>$ or replacing the tail with $t'_c \neq t_c$ results in $<h_c, r_c, t'_c, w_c>$.

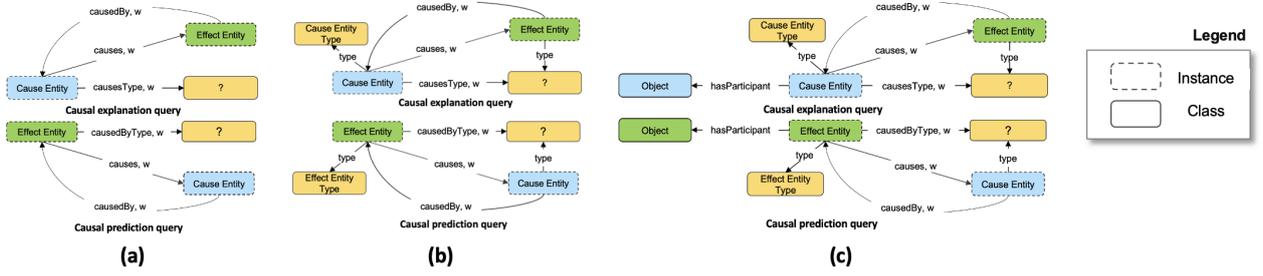

Figure 6: Different CLEVRER-Humans CausalKG structures used for evaluating causal explanation and prediction tasks: (a) subgraph C which consists of links with only causal relations, i.e. *causes, causedBy, causesType,* and *causedByType,* (b) subgraph CT with causal relations and information about entity types, i.e. *rdf:type,* (c) subgraph CTP with causal relations, entity type relations, and information about the objects that participate in the causal events.

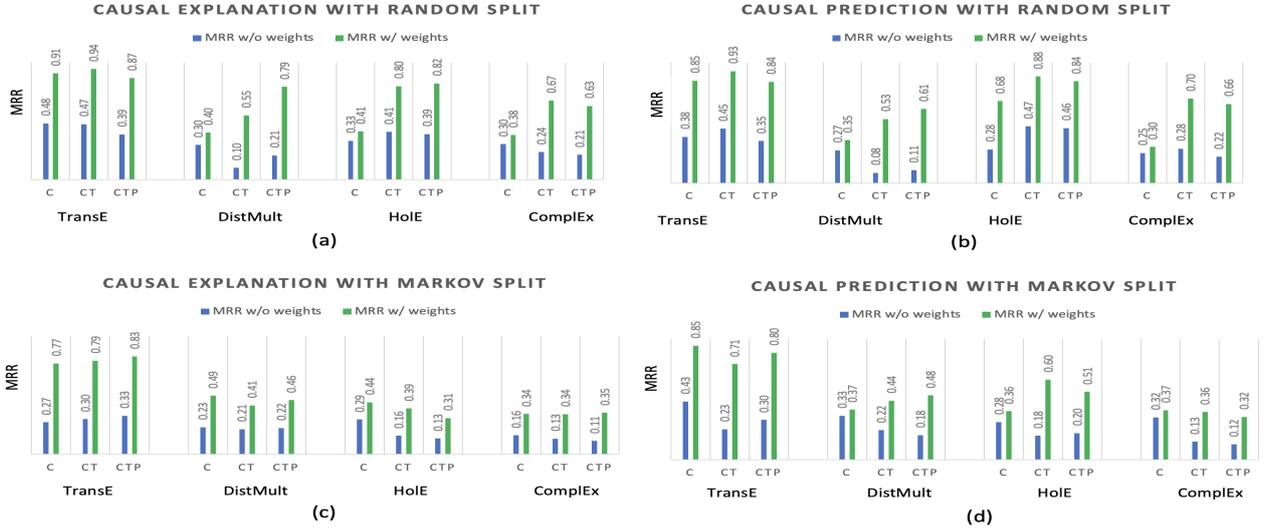

Figure 7: Evaluation of KGE models without causal weights (i.e. CausalKGE-Base) and with causal weights (i.e., CausalKGE-W) using different CausalKG structures and data split strategies (i.e., random-data split or Markov-based data split): (a) CausalKGE-Base vs CausalKGE-W for causal explanation with random data split, (b) CausalKGE-Base vs CausalKGE-W for causal prediction with random data split, (c) CausalKGE-Base vs CausalKGE-W for causal explanation with Markov-based data split, and (d) CausalKGE-Base vs CausalKGE-W for causal prediction with Markov-based data split.

The model scores the true link $<h_c, r_c, t_c, w_c>$ and corrupted links $<h'_c, r_c, t_c, w_c>$, $<h_c, r_c, t'_c, w_c> \in \mathcal{T}'$. The scores are then sorted to obtain the rank of the true link. The filtered evaluation setting and filtered corrupted links $\mathcal{T}'$ are used to exclude the links present in the training and validation set. The overall performance of the models is measured using mean reciprocal rank (MRR) and Hits@K for k = {1,3,10}. MRR is the mean over the reciprocal of individual ranks of test links. Hits@k is the ratio of test links present among the top k ranked links.

## 6 RESULT AND DISCUSSION

CausalLP is evaluated for the causal explanation and prediction task, using the CausalKG generated from the CLEVRER-Humans dataset. Specifically, CasualLP has four trained KGE models:

- CausalKGE-Base with random data split
- CausalKGE-Base with Markov-based data split
- CausalKGE-W with random data split
- CasualKGE-W with Markov-based data split

Figure 7 and Figure 8 show the MRR scores for the four KGE models evaluated on different CausalKG structures, i.e., C, CT and CTP. The MRR scores of CausalKGE-W for causal explanation and causal prediction using the random data split, on average across KGE models, outperforms CausalKGE-Base by 43.26% and 79.26% respectively.

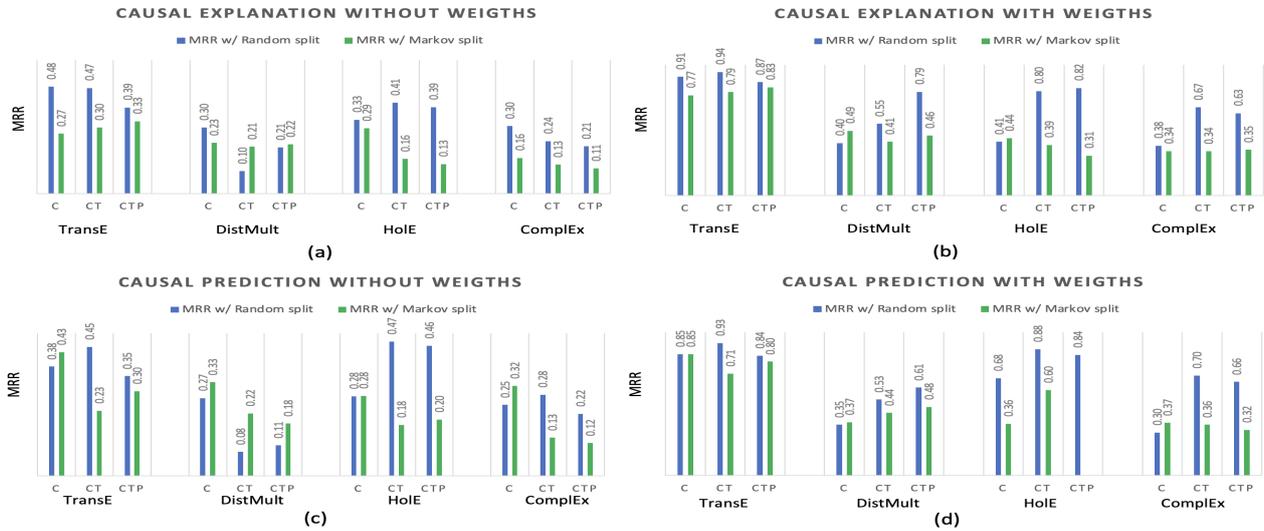

Figure 8: Evaluation of KGE models without causal weights (i.e., CausalKGE-Base) and with causal weights (i.e., CausalKGE-W) using different CausalKG structures and data split strategies (i.e., random-data split or Markov-based data split): (a) CausalKGE-Base for causal explanation with random vs Markov-based data split, (b) CausalKGE-W for causal explanation with random vs Markov-based data split, (c) CausalKGE-Base for causal prediction with random vs Markov-based data split, and (c) CausalKGE-Base for causal prediction with random vs Markov-based data split.

The MRR scores of CausalKGE-W for causal explanation and causal prediction using Markov-based data split, on average across KGE models, outperforms CausalKGE-Base by 115.05% and 38.96% respectively. The MRR scores of CausalKGE-W using random-based data split when enriched with additional knowledge for causal explanation,i.e., CTP, on average across KGE models, outperforms C by 33.49%. The MRR scores of CausalKGE-W using Markov-based data split for causal explanation, on average across KGE models, outperforms random data split by 0.75%. The MRR scores of CausalKGE-Base using Markov-based data split for causal prediction, on average across KGE models, outperforms random data split by 15.28%. The MRR scores of CausalKGE-W using random-based data split when enriched with additional knowledge for causal prediction,i.e., CTP, on average across KGE models, outperforms C by 28.65%. Along with the MRR score, we also estimated Hit@k for K. The Hit@1, Hit@3 and Hit@10 of CausalKGE-W for causal explanation using random based split outperformed CausalKGE-Base by 37.28%, 31.10%and 68.2%, respectively. The Hit@1, Hit@3 and Hit@10 of CausalKGE-W for causal prediction using random based split outperformed CausalKGE-Base by 84.22%, 64.62%, and 80.57% respectively. The Hit@1, Hit@3 and Hit@10 of CausalKGE-W for causal prediction using Markov-based split outperformed CausalKGE-Base by 29.91%, 34.33%, and 36% respectively. The Hit@1, and Hit@10 of CausalKGE-W for causal explanation using Markov-based split outperformed CausalKGE-Base by 145.65% and 114.38% respectively.

The evaluation results show improved performance of CausalLP for causal prediction and causal explanation using CausalKGE-W. However, we observed CausalLP for random-data split outperformed the Markov-based data split for both CausalKGE-Base and CausalKGE-W due to issues of data leakage and model bias [1].

We also noticed adding knowledge to the CausalKG structure, the CausalKGE-W significantly outperform CausalKGE-Base for both random based and markov based data split.

We restricted our KGE models to TransE, DistMult, ComplEx, HolE due to the implementation limitation of FocusE. FocusE is implemented in ampligraph. The ampligraph currently supports only the above KGE models. However, in the future, we would like to evaluate and compare FocusE with other KGE models as well.

## 7 CONCLUSION

In this paper, we introduced CausalLP, an approach to finding missing relations in an incomplete causal network using knowledge graph link prediction. The approach addresses a crucial gap in the state-of-the-art by considering causal weights along with a causal links. The KGE models trained with causal weights outperform all baseline KGE metrics without causal weights. The results demonstrate that an effective fusion of causal links with causal weights in a KG can facilitate the completion of sparse KGs that may be missing critical causal relations. In the future, the proposed approach will be extended with a more varied selection of KG embedding models.